\documentclass[fleqn]{llncs}
\usepackage{amsmath,amssymb,pstricks,pst-node}
\newcommand{\rangebopd}[5]	{#1_#2#4 #5 #4#1_#3}
\newcommand{\rangeb}[3]		{\rangebopd{#1}{#2}{#3},{\ldots}}
\newcommand{\rangenop}[2]	{\rangebopd{#1}1n{#2}{\ldots}}
\newcommand{\rangen}[1]		{\rangebopd{#1}1n,{\ldots}}
\newcommand{\rangecbop}[4]	{\rangebopd{#1}{#2}{#3}{#4}{\cdots}}

\newcommand{\setc}[2]		{\left\{\,{#1}\mid{#2}\,\right\}}
\newcommand{\tupleII}[2]	{\left\langle{#1};{#2}\right\rangle}

\newcommand{\mi}[1]		{\mathit{#1}}
\newcommand{\ms}[1]		{\mathsf{#1}}
\newcommand{\mr}[1]		{\mathrm{#1}}

\newcommand{\ie}		{i.e.}
\newcommand{\eg}		{e.g.}

\newcommand{\qtext}[1]		{\quad\text{#1}\quad}
\newcommand{\qqtext}[1]		{\qquad\text{#1}\qquad}

\newcommand{\eclipse}		{\textup{ECL\textsuperscript{\textit{i}}PS%
				\textsuperscript{\textit{e}}}}

\newcommand{\Regions}		{\mi{Regions}}
\newcommand{\RCC}		{\ms{RCC8}}
\newcommand{\Rel}		{\mi{Rel}}
\newcommand{\C}			{\mathcal{C}}
\newcommand{\D}			{\mathcal{D}}

\begin{document}

\title{Relation Variables in \\Qualitative Spatial Reasoning}
\author{Sebastian Brand}
\institute{National Research Institute for Mathematics and Computer Science (CWI)\\
	P.O.\;Box 94079, 1090 GB, Amsterdam, The Netherlands}

\maketitle

\begin{abstract}
We study an alternative to the prevailing approach to modelling qualitative
spatial reasoning (QSR) problems as constraint satisfaction problems.
In the standard approach, a relation between objects is a constraint
whereas in the alternative approach it is a variable.

By being declarative, the relation-variable approach greatly
simplifies integration and implementation of QSR.
To substantiate this point, we discuss several specific QSR algorithms
from the literature which
in the relation-variable approach reduce to the customary constraint propagation
algorithm enforcing generalised arc-consistency.
\end{abstract}

%=====================================================================

\section{Introduction}

Qualitative spatial representation and reasoning (QSR)
\cite{cohn:2001:qualitative}
lends itself well to modelling by constraints.
In the standard approach,
a spatial object, such as a region, is described by a variable,
and the qualitative relation between spatial objects,
such as a topological relation between two regions, contributes a constraint.
For many QSR calculi, it is known that
if all the constraints represent definite (base) relations
and path-consistency (PC) holds,
then this description of a spatial scene is consistent.
If the relation is not fully specified,
the corresponding constraint is a disjunction of basic constraints.
By establishing PC, such a disjunctive constraint is refined in view of the
constraints with which it shares a variable.
A combination of PC with search over the disjunctive constraints
decides the consistency of indefinite scene descriptions.

We examine here an alternative constraint-based formulation of QSR.
In this approach, a spatial object is a constant,
and the relation between spatial objects is a variable.
We call this the \emph{relation-variable} approach,
in contrast to the conventional \emph{relation-constraint} approach above.
Although modelling QSR with relation variables
is not original, see \cite{tsang:1987:consistent},
it is mentioned very rarely.
This fact surprises in view of the advantages of this approach.
In particular, the following two important issues are tackled successfully:
\begin{description}
\item[Integration.]
	Space has several aspects that can be characterised
	qualitatively, such as size, shape, orientation.
	These aspects are interdependent,
	but no convenient canonical representation exists
	that provides a link
	(the role of time points in temporal reasoning).
	Spatial reasoning problems in practice are also not likely to occur
	in pure form. They may be embedded into a non-spatial context,
	or contain application-specific side constraints.

	The relation-variable approach to QSR is declarative in a strict sense
	and is thus well-suited for these integration problems.
\item[Systems.]
	Typical current constraint solving platforms
	focus on domain reduction, and accordingly
	provide convenient access to variable domains.
	Modifying the constraint network, on the other hand,
	is usually difficult.
	This task is, however, required for enforcing PC.

	A formulation of QSR according to the relation-variable approach
	means that generic domain-reducing propagation algorithms and
	conventional constraint solving platforms
	can be used instead of dedicated spatial reasoning systems.
\end{description}

%---------------------------------------------------------------------

\paragraph{Plan of the paper.}

We begin by introducing briefly the
necessary constraint solving concepts and methods,
and qualitative spatial reasoning,
using the example of the RCC-8 calculus.
The next section presents in-depth the two modelling approaches
for constraint-based QSR.
In the following sections, we discuss several aspects of space
and contrast the relation-variable and relation-domain approach.
We finally mention some new modelling options,
and end with a summary.

%---------------------------------------------------------------------

\subsection{Constraint Satisfaction}

Recent coverage of the field can be found in
\cite{apt:2003:principles,dechter:2003:constraint,fruehwirth:2003:essentials}.

Consider a sequence  $X = \rangeb x1m$ of pairwise different variables
with respective domains $\rangeb D1m$.
By a \emph{constraint} $C$ on $X$, written $C(X)$,
we mean a subset of $\rangecbop D1m{\times}$.
The arity of $C$ is $m$.
A \emph{constraint satisfaction problem (CSP)} consists of a finite sequence
of variables $X= \rangen x$ with respective domains $\D = \rangen D$,
and
a finite set $\C$ of constraints, each on a subsequence of $X$\@.
We write it as $\tupleII{\C}{x_1 \in D_1, \ldots, x_n \in D_n}$,
or shorter as $\tupleII{\C}{X \in \D}$.
Given an element $d = \rangen d$ of $\rangenop D{\times}$
and a subsequence \mbox{$Y = \rangeb x{{i_1}}{{i_\ell}}$} of $X$
we denote by $d[Y]$ the sequence $\rangeb d{{i_1}}{{i_\ell}}$;
in particular, we have $d[x_{k}] = d_{k}$.
A \emph{solution} to $\tupleII{\C}{X \in \D}$ is an element $d \in \D$
such that for each constraint $C \in \C$ on the variables $Y$ we have $d[Y] \in C$.

%---------------------------------------------------------------------

\subsubsection{Constraint propagation.}

One method to
establish satisfiability of CSPs
when the search space is finite
is systematic search for a solution.
For reducing the search space and overall search effort,
constraint propagation is often very useful;
the principle is to replace a given CSP by another one
that is equivalent with respect to the solutions but
that is easier to solve.
Constraint propagation is typically characterised
by the resulting \emph{local consistency}.
The two notions most relevant for this paper are:
\begin{description}
\item[\emph{Path Consistency (PC)}:]
	A CSP of binary constraints is path-consistent
	\cite{montanari:1974:networks}
	if for every triple of variables $x,y,z$
	\[
	C(x,z) = \setc{ (a,c) }{ b \ \text{exists s.t.}\ (a,b) \in C(x,y)
		\ \text{and}\ (b,c) \in C(y,z) }.
	\]
	It is assumed here that a unique constraint $C(u,w)$ for each pair
	of variables $u,w$ exists, and that $C(u,w) = C^{-1}(w,u)$.

\item[\emph{Generalised Arc-Consistency (GAC)}:]
	A constraint $C(X)$ is generalised arc-consistent
	\cite{mohr:1988:good}
	if for all $x_k \in X$ and all $a \in D_k$
	\[
	d \in C(X) \quad\text{exists such that}\quad d[x_k]=a.
	\]
	In short, every domain value must participate in a local solution.

	A CSP is generalised arc-consistent if each of its constraints is.

	\smallskip
	For example, the CSP $\tupleII{x + y = z}{\; x,y,z \in \{1,2,3\}}$
	can be reduced to $\tupleII{x + y = z}{\; x,y \in \{1,2\}, z \in \{2,3\}}$
	which is GAC.
\end{description}
Enforcing PC means reducing constraints but not domains, whereas
enforcing GAC means reducing domains but not constraints.

A number of generic methods to establish GAC for a constraint are known,
and many constraint solving systems have implementations.
One example is the \emph{GAC-schema} \cite{bessiere:1997:arc-consistency}
available in ILOG Solver \cite{ilog:2001:solver}.

%---------------------------------------------------------------------

\subsection{Qualitative Spatial Reasoning}

\begin{figure}
\center
\psset{nodesep=0.2, linewidth=0.3pt, unit=9mm}
\begin{pspicture*}(0.5,1)(11.6,5)

\pscircle(1,3.7){0.5}	\rput{U}(1,3.7){a}
\pscircle(1,2.3){0.5}	\rput{U}(1,2.3){b}
\pnode(1.3,3){disjoint}
\rput{U}(1,1.4){$\ms{DC}$}

\pnode(2.6,3){meet1}
\pscircle(3,3.5){0.5}	\rput{U}(3,3.5){a}
\pscircle(3,2.5){0.5}	\rput{U}(3,2.5){b}
\pnode(3.4,3){meet2}
\rput{U}(3,1.6){$\ms{EC}$}

\pnode(4.5,3){overlap1}
\pscircle(5,3.3){0.5}	\rput{U}(5,3.5){a}
\pscircle(5,2.7){0.5}	\rput{U}(5,2.5){b}
\pnode(5.5,3){overlap2}
\rput{U}(5,1.8){$\ms{PO}$}

\cnode(7,3){0.5}{equal}		\rput{U}(6.8,3.1){a}
\rput[l]{U}(6.6,2){$\ms{EQ}$}		\rput{U}(7.2,2.9){b}

\cnode(9,4){0.5}{coveredby}	\rput{U}(8.8,4){a}
\pscircle(8.8,4){0.3}		\rput{U}(9.3,4){b}
\rput{U}(9,4.8){$\ms{TPP}$}

\cnode(9,2){0.5}{covers}	\rput{U}(8.7,2){a}
\pscircle(9.2,2){0.3}		\rput{U}(9.2,2){b}
\rput{U}(9,1.2){$\ms{TPPi}$}

\cnode(11,4){0.5}{inside}	\rput{U}(10.9,4){a}
\pscircle(10.9,4){0.3}		\rput{U}(11.35,4){b}
\rput{U}(11,4.8){$\ms{NTPP}$}

\cnode(11,2){0.5}{contains}	\rput{U}(10.65,2){a}
\pscircle(11.1,2){0.3}		\rput{U}(11.1,2){b}
\rput{U}(11,1.2){$\ms{NTPPi}$}
\end{pspicture*}
\caption{RCC-8 relations (2D example)}
\label{figure:rcc8}
\end{figure}
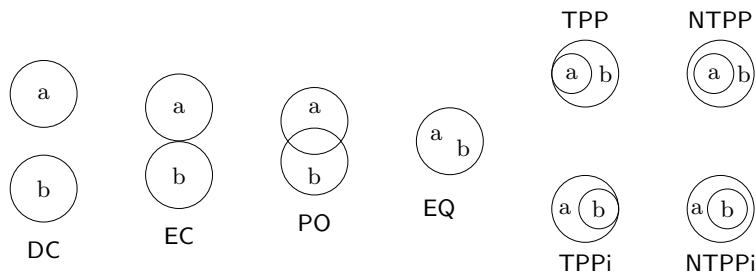

The topological calculus RCC-8 \cite{randell:1992:spatial}
is one of the best-known formalisations in spatial reasoning.
We use it to illustrate a number of concepts.
In \mbox{RCC-8} one distinguishes 8 topological relations
between two regions, see Fig.~\ref{figure:rcc8}:
\emph{disconnected, externally connected, partially overlapping, equal,
tangential proper part, non-tangential proper part},
and inverses of the latter two.
These are denoted
$\ms{DC}, \ms{EC}, \ms{PO}, \ms{EQ}, \ms{TPP}, \ms{NTPP}, \ms{TPPi}, \ms{NTPPi}$,
respectively;
together they form a set that we call $\RCC$.

\paragraph{Jointly exhaustive and pairwise disjoint.}
Any two spatial regions are in one and exactly one of the RCC-8
relations to each other.

\paragraph{Composition table.}
Considering the triple $R_{a,b}, R_{b,c}, R_{a,c}$
of relations between regions $a,b,c$,
one finds that not all triples of RCC-8 relations
are semantically feasible.
The consistent triples are collected in the RCC-8 composition table.
It contains 193 relation triples, such as $(\ms{NTPP}, \ms{EC}, \ms{DC})$.
Bennett \cite{bennett:1998:determining} proved that
compositional consistency entails global consistency:
if for all triples of regions the relations between them respect
the composition table then this topological scenario is consistent.

\paragraph{Converse relation table.}
In analogy to the composition table,
it is helpful to think of a converse relation table consisting
of the 8 pairs $(R,Ri)$ of RCC-8 relations such that
$Ri$ is the converse of $R$. It contains for example
$(\ms{EQ}, \ms{EQ})$ and $(\ms{TPP}, \ms{TPPi})$.
If we agree on $(\ms{EQ})$ for the relation of a region with itself
then the converse relation table follows from the composition table.

%=====================================================================

\section{Approaches to Constraint-based QSR}

A spatial topological scenario consists of
a set of region names denoted by $\Regions$,
and possibly some restrictions on the topological relation for regions pairs.
A scenario is fully specified if for each region pair
exactly one RCC-8 relation is given.

We examine now how scenarios can be modelled as constraint
satisfaction problems.
We continue using topology with RCC-8 as an example,
but most of the concepts below are immediately transferable
to other spatial aspects.

%---------------------------------------------------------------------

\subsection{Relations as Constraints}

In this conventional approach, $\Regions$ is considered to be
a set of region variables.
Their infinite domain is the set of all
spatial regions in the underlying topological space;
for example, if we model 2D space then a region variable
represents a set of points in the plane.
Information about the topological relation between two regions
is expressed as a binary constraint $\Rel$ that
corresponds to a subset of $\RCC$.
One usually writes this in infix notation as
\[
	\text{constraint}\quad
	x \mathbin{\Rel} y
	\quad
	\qtext{where}
	\Rel \subseteq \RCC
	\qtext{and}
	x,y \in \Regions.
\]
Such a CSP describes a possibly partially specified scenario.
Whether a corresponding fully specified and satisfiable scenario
exists is checked by path-consistency
and search over the relations.
A PC-enforcing algorithm revises the constraints between regions according
to the converse relation and composition tables of RCC-8,
and search branches over disjunctive constraints.

Establishing satisfiability of a scenario processes only the constraints,
for compositional consistency.
The variables remain unassigned.

%---------------------------------------------------------------------

\subsection{Relations as Variables}\label{section:relvar-approach}

Here we interpret every element of $\Regions$ as a constant.
The topological relation between two regions is a variable
with a subset of $\RCC$ as its domain.
Such a relation variable exists
for each ordered pair of regions,
and we collect all these variables in an array $\Rel$.
We write an individual relation as
\[
	\text{variable}\quad
	\Rel[a,b]
	\quad
	\qtext{where}
	\Rel[a,b] \subseteq \RCC
	\qtext{and}
	a,b \in \Regions.
\]

%---------------------------------------------------------------------

\subsubsection{Integrity constraints.}

Relation converse and composition in this setting
are captured at the constraint level.
The binary constraint $\ms{conv}$ represents the converse relation table:
\[
	\ms{conv}(\, \Rel[a,b],\; \Rel[b,a] \,)
	\quad\qtext{for all} \{a,b\} \subseteq \Regions.
\]
The composition table is represented by the ternary constraint $\ms{comp}$, with
\[
	\ms{comp}(\,\Rel[a,b],\; \Rel[b,c], \; \Rel[a,c]\,)
	\qquad\text{for all}\quad \{a,b,c\} \subseteq \Regions.
\]
In presence of
\[
	\Rel[a,a] = \ms{EQ} \quad\qtext{for all} a \in \Regions
\]
and a $\ms{conv}$ constraint for all pairs of different regions,
one $\ms{comp}$ constraint per three different regions
suffices.

%---------------------------------------------------------------------

\subsection{Comments}

By modelling the items of interest as variables and
static information as constraints,
the relation-variable approach yields plain finite-domain CSPs
in which the solutions (\ie, assignments) are relevant.
There is a straightforward correspondence between
a solution and a fully specified, consistent scenario.
Obtaining the latter from a partially specified scenario
amounts to the standard task of solving a finite-domain CSP.

Constructing a relation-variable model
means finding integrity constraints
that embody the intended semantics.
Once that has been established,
the origin or meaning of the constraints is irrelevant.
For example, a constraint solver can ignore whether
$\ms{comp}$ represents the composition operation in a relation algebra;
we also discuss examples below in which other restrictions on the relations
must be satisfied.
There is thus a clear distinction between specification and execution.
The relation-variable approach is declarative in a strict sense.

\subsubsection{Constraint propagation.}

The relation-variable approach is independent of the particular
constraint solving method.  We could, however, choose
a solver based on search and propagation, and furthermore we could choose
a GAC-enforcing propagation algorithm.

Path consistency in the relation-constraint approach
and
generalised arc-consistency in the relation-variable approach
simulate each other.
This can be seen by analysing, in both approaches,
the removal of one topological relation
from the disjunctive constraint $a \mathbin{\Rel} b$,
or from the domain of the variable $\Rel[a,b]$, respectively.
The reason in both cases must be the lack
of supporting relations between $a,c$ and $b,c$,
for some third region $c$; that is, compositional consistency.

\subsubsection{Complexity.}

It is perhaps not surprising but useful to mention
that establishing the respective local consistency in either approach
(\ie, PC and GAC) requires the same computational effort.
Let $n$ denote the number of regions.
Enforcing PC by an algorithm as the one given in
\cite{mackworth:1977:consistency} requires time in
$O(n^3)$ \cite{mackworth:1985:complexity}.
For this, one assumes that one PC step,
restricting $a \mathbin{\Rel} c$ by
$a \mathbin{\Rel} b$ and $b \mathbin{\Rel} c$, takes constant time.

Analogue reasoning entails that GAC can be enforced in constant time
on a single $\ms{comp}( \Rel[a,b], \Rel[b,c],  \Rel[a,c] )$ constraint ---
observe that the three variables have domains of size at most eight.
In this way, the overall time complexity depends only on the number of such
constraints, and is thus in $O(n^3)$.

\subsubsection{Previous work.}

Tsang \cite{tsang:1987:consistent} describes
the relation-variable approach
in qualitative temporal reasoning,
a field similar to QSR.
The idea appears not to have caught on, however.
One reason is probably that integration in temporal reasoning
is simpler because the canonical representation of time points
on the real line exists.
By referring to its end points,
a time interval can directly be related to its
duration or another time interval.
Space, in contrast, has no such convenient
canonical representation --- but many aspects to be integrated.

In QSR, the possibility of the relation-variable approach is mentioned
occasionally in passing, but without examining its potential.
For actually modelling and solving QSR problems
using relation variables I am only aware of
\cite[pages 30-33]{apt:2003:principles},
which deals with a single aspect (topology) only.

%=====================================================================

\section{Relation Variables in Use}
\label{section:integration}

An essential advantage of the relation-variable approach is
that the relevant information is available in variables.
This means that linking pieces of information
reduces to merely stating additional constraints on the variables.
In that way, embedding a QSR problem into an application context
or adding side restrictions,
for example, can be dealt with easily and declaratively.

We illustrate the issue of composite models with the case of aspect integration.

%---------------------------------------------------------------------

\subsection{Combining Topology and Size}
\label{section:topology-size}

Following Gerevini and Renz \cite{gerevini:2002:combining},
we study scenarios combining topological and size information.
We collect information about both these aspects and their link in one CSP.

Let $n$ be the number of regions.

\subsubsection{Topological aspect.}
As in Section~\ref{section:relvar-approach}, the
\[
	n \times n \text{ array } \mi{TopoRel}
\]
of RCC-8 relation variables stores the topological relation
between two regions.
The integrity constraints $\ms{conv}_{\mr{RCC8}}$, $\ms{comp}_{\mr{RCC8}}$
need to hold.

%---------------------------------------------------------------------

\subsubsection{Size aspect.}
Relative size of regions is captured by one of $\{<, =, >\}$,
as in \cite{gerevini:2002:combining}.
The
\[
	n \times n \text{ array } \mi{SizeRel}
\]
of variables stores the relative sizes of region pairs.
The converse relation and composition tables
are straightforward; the integrity constraints are
\begin{align*}
	\ms{conv}_{\mr{Size}} &= \{\, (<, >),\; (=, =),\; (>, <) \,\},
	\qquad\text{and}\\
	\ms{comp}_{\mr{Size}} &= \{\, (<, <, <),\; (<, =, <),\; \ldots \}
	\qquad\text{(13 triples).}
\end{align*}

%---------------------------------------------------------------------

\subsubsection{Linking the aspects.}

The topological relation between two regions is
dependent on their relative size.
A table with this information is given in \cite{gerevini:2002:combining},
it contains rules such as the following:
\[\begin{array}{lcl}
	\mi{TopoRel}[x,y] = \ms{TPP}
	&\qqtext{implies}&
	\mi{SizeRel}[x,y] = (<),
	\\
	\mi{SizeRel}[x,y] = (=)
	&\qqtext{implies}&
	\mi{TopoRel}[x,y] \in \{\ms{DC}, \ms{EC}, \ms{PO}, \ms{EQ}\}.
\end{array}\]
In \cite{gerevini:2002:combining}, these rules represent a meta constraint.
Here, we infer the linking constraint
\[
	\ms{link}_{\mr{Topo\&Size}} = \{\, (\ms{TPP}, <),\; (\ms{DC}, =),\; \ldots \}
	\qquad\text{(14 pairs)}
\]
which is to be stated as
\[	\ms{link}_{\mr{Topo\&Size}}(\, \mi{TopoRel}[a,b],\; \mi{SizeRel}[a,b] \,)
\]
for all regions $a,b$.

%---------------------------------------------------------------------

\paragraph{Example.}

Let us pick up the combined scenario from \cite[p. 14]{gerevini:2002:combining}.
Five regions, denoted by $\{0,\ldots,4\}$, are constrained by
\begin{align*}
	&\mi{TopoRel}[0,2] \in \{\ms{TPP}, \ms{EQ}\}
	&&\mi{SizeRel}[0,2] \in \{<\}		\\
	&\mi{TopoRel}[1,0] \in \{\ms{TPP}, \ms{EQ}, \ms{PO}\}
	&&\mi{SizeRel}[3,1] \in \{<,=\}		\\
	&\mi{TopoRel}[1,2] \in \{\ms{TPP}, \ms{EQ}\}
	&&\mi{SizeRel}[2,4] \in \{<,=\}		\\
	&\mi{TopoRel}[4,3] \in \{\ms{TPP}, \ms{EQ}\}
\end{align*}
Independently, the topological and the size scenarios are consistent
while the combined scenario is not.
It is pointed out in \cite{gerevini:2002:combining}
that naive propagation scheduling schemes
do not suffice to detect inconsistency.

A formulation of this scenario as a combined topological \& size CSP
in the relation-variable approach is straightforward.
The resulting CSP can be entered into a constraint programming platform
such as {\eclipse} \cite{wallace:1997:eclipse}.
{\eclipse} is focused on search and domain-reducing propagation;
in particular, it offers a GAC-enforcing propagation algorithm
for user-defined constraints.
Given our CSP in {\eclipse}, solely executing GAC-propagation for all constraints
yields failure, which proves that this CSP is inconsistent.
\qed

\bigskip

For the same purpose
but within the relation-constraint approach,
Gerevini and Renz proposed a new algorithm
called \textsc{Bipath-consistency} \cite{gerevini:2002:combining}.
Its principle is the computation of path-consistency for both types of relations
in an interleaved fashion while taking into account the interdependency.
The $\ms{link}_{\mr{Topo\&Size}}$ constraint is
in essence treated as a \emph{meta constraint} on the algorithm level.
Moreover, the \textsc{Bipath-consistency} algorithm
fixes in part the order of propagation.

The relation-variable method, on the other hand, is declarative;
all information is in the five types of constraints.
They are handled by repeated, interleaved calls to the same GAC-enforcing algorithm.
The actual propagation order is irrelevant for the result.

\textsc{Bipath-consistency} is restricted to combining
two types of relations (\eg, two aspects of space).
In contrast, the relation-variable approach is compositional in the sense
that adding a third aspect, such as morphology \cite{cristani:1999:complexity}
or orientation, is straightforward.
It amounts to formulating integrity constraints
(\eg, $\ms{conv, comp}$), linking constraints
to each of the already present aspects,
and a constraint linking all three aspects.
Some of these constraints may be logically redundant.

%---------------------------------------------------------------------

\subsection{Combining Cardinal Directions and Topology}
\label{section:cardinal-directions}

In orientation, another important aspect of space,
one studies the relation of two objects, the primary and the reference object,
with respect to a frame of reference.
It is thus inherently a ternary relation,
but by agreeing on the frame of reference, a binary relation is obtained.

The binary relation approach is realised in the cardinal direction model
\cite{frank:1992:qualitative}, based on the geographic (compass) directions.
Points as well as regions have been studied as the objects to be oriented.
The point-based models can be cast in the relation-variable approach
analogously to topology, Section~\ref{section:relvar-approach}.
For instance, Frank~\cite{frank:1992:qualitative}
distinguishes the jointly exhaustive and pairwise disjoint
relations $\ms{N}, \ms{NW}, \ms{W}, \ldots$
for points; denoting North, Northwest, West, and so on.
Ligozat~\cite{ligozat:1998:reasoning} gives a composition table.

\subsubsection{Orienting regions.}
Goyal and Egenhofer~\cite{goyal:1997:direction}
and Skiadopoulos and Koubarakis~\cite{skiadopoulos:2001:composing}
study a more expressive model, in which the oriented objects are regions.
The exact shape of the primary region is taken into account,
and a ninth atomic relation $\ms{B}$ exists,
describing overlap of the primary region and
the axes-parallel minimum bounding box of the reference region.
\emph{Sets} of the atomic relations are then used to describe
directional information.
In this way, for example,
the position of South America
for an observer located in Ecuador
can be fully described by the set
$\{\ms{B}, \ms{N}, \ms{NE}, \ms{E}, \ms{SE}, \ms{S}\}$.
In contrast, the position of Ecuador with respect to South America
is just $\{\ms{B}\}$.

Relation variables for directional information
are thus naturally \emph{set variables}:
they take their value from a set of sets of constants,
unlike relation variables for topology and size
whose domain is a set of atomic constants.

For each pair $a,b$
of regions, the direction is a relation variable
\[
	\mi{DirRel}[a,b] \in \mathcal{P}(\ms{Dir})
	\qquad
	\text{where}\quad
	\ms{Dir} = \{\ms{B}, \ms{N}, \ms{NW}, \ldots, \ms{NE}\}.
\]
$\mathcal{P}$ denotes the power set function.

%---------------------------------------------------------------------

\subsubsection{Integrity constraints.}

A restriction on the
set values that $\mi{DirRel}[a,b]$ can take
arises if $a,b$ are internally connected regions, which is often assumed.
Only 218 of the 512 subsets of $\ms{Dir}$ are
then semantically possible.
This knowledge can be represented in a unary integrity constraint,
which for example
allows $\{\ms{N}, \ms{NE}, \ms{E}\}$
but
excludes $\{\ms{N}, \ms{S}\}$.
The usual integrity constraints $\ms{comp}$ and $\ms{conv}$ can be derived
from studies of composition \cite{skiadopoulos:2001:composing}
and converse \cite{cicerone:2004:cardinal}
(but it is outside of our focus whether these 
are the only integrity constraints needed).

%---------------------------------------------------------------------

\subsubsection{Integration with topology.}

Let us briefly consider linking directional information to topology.
The relevant knowledge could be expressed by rules as
\[\begin{array}{lcl}
	\mi{TopoRel}[x,y] \in \{\ms{EQ},\ms{NTPP},\ms{TPP}\}
	&\qqtext{implies}&
	\mi{DirRel}[x,y] = \{\ms{B}\},
	\\
	\mi{TopoRel}[x,y] \in \{\ms{NTPPi},\ms{TPPi}\}
	&\qtext{implies}&
	\mi{DirRel}[x,y] \supseteq \{\ms{B}\},\\
\end{array}\]
from which a constraint $\ms{link}_{\mr{Topo\&Dir}}$
can be defined.  It is to be stated as
\[
	\ms{link}_{\mr{Topo\&Dir}}(\, \mi{TopoRel}[a,b],\; \mi{DirRel}[a,b] \,)
\]
for all regions $a,b$.
We now have some components of a combined cardinal directions \& topology
model.  It can be given to any sufficiently expressive constraint solver,
which in particular would provide constraints on set variables.

Constraint solving with set variables is discussed
in \cite{gervet:1997:interval}.
Many contemporary constraint programming systems support
set variables.

%---------------------------------------------------------------------

\subsection{Cyclic Ordering of Orientations with Relation Variables}

\begin{figure}
\centering
\psset{unit=4mm}
\begin{pspicture*}(0,0)(25,7)
\psset{linewidth=1pt,arrows=<-,arrowsize=0.3}

\psline[linewidth=1.2pt](1,6)(1,0)
\rput(0.5,5.2){a}
\rput(1.5,4.5){b}

{\psset{origin={-8,-3}}
\SpecialCoor
\psline(3;90)(3;270)
\psline(3;50)(3;230)
\NormalCoor
}%
\rput(7.3,5.3){a}
\rput(9.8,4){b}

\psline[linewidth=1.2pt,arrows=<->](15,6)(15,0)
\rput(14.5,5.2){a}
\rput(15.5,0.8){b}

{\psset{origin={-22,-3}}
\SpecialCoor
\psline(3;90)(3;270)
\psline(3;200)(3;20)
\NormalCoor
}%
\rput(21.5,5.2){a}
\rput(20,3){b}
\end{pspicture*}
\caption{The relations $\ms{e,l,o,r}$ of a pair of orientations}
\label{fig:relvars:orientations}
\end{figure}
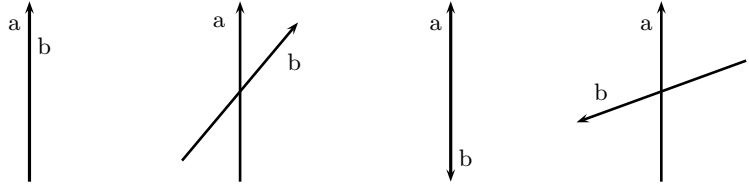

\begin{figure}
\centering
\psset{unit=4mm}
\begin{pspicture*}(0,0)(22,7)
\psset{linewidth=1pt,arrows=<-,arrowsize=0.3}

{\psset{origin={-3,-3}}
\SpecialCoor
\psline(3;90)(3;270)
\psline(3;135)(3;-45)
\psline(3;200)(3;20)
\NormalCoor
}%
\rput(3.8,5.5){a}
\rput(1,1.5){b}
\rput(1,4){c}

{\psset{origin={-11,-3}}
\SpecialCoor
\psline(3;90)(3;270)
\psline[linewidth=1.2pt](3;135)(3;-45)
\NormalCoor
}%
\rput(11.8,5.5){a}
\rput(10,4.7){b}
\rput(9,4){c}

{\psset{origin={-19,-3}}
\SpecialCoor
\psline(3;90)(3;270)
\psline[linewidth=1.2pt,arrows=<->](3;230)(3;50)
\NormalCoor
}%
\rput(19.7,5.5){a}
\rput(17.2,2){b}
\rput(21,4.5){c}
\end{pspicture*}
\caption{The $\ms{Cyc}$ relations $\ms{lrl}$, $\ms{lel}$,
	and $\ms{rol}$ of a triple of orientations}
\label{fig:relvars:cyc}
\end{figure}
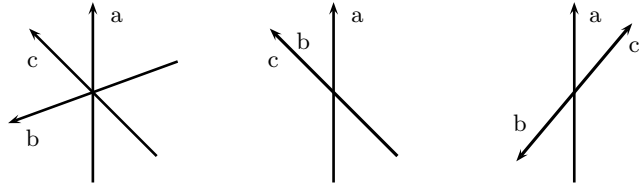

From the several formalisations of orientation information
with an explicit frame of reference,
let us examine the approach of Isli and Cohn
to cyclic ordering of 2D orientations
\cite{isli:2000:new}.
Here, the spatial objects are orientations, \ie\ directed lines.
At the root of the framework is the qualitative classification
of the angle $\alpha = \sphericalangle(a, b)$ between the two orientations $a$ and $b$ by
\[
	\ms{Or}(\alpha) =
	\begin{cases}
	\ms{e} \qtext{(equal)} &\text{if}\quad \alpha = 0,\\
	\ms{l} \qtext{(left)} &\text{if}\quad 0 < \alpha < \pi,\\
	\ms{o} \qtext{(opposite)} &\text{if}\quad \alpha = \pi,\\
	\ms{r} \qtext{(right)} &\text{if}\quad \pi < \alpha < 2\pi
	\end{cases}
\]
into the jointly exhaustive and pairwise disjoint relations
$\ms{e},\ms{l},\ms{o},\ms{r}$.
See Fig.~\ref{fig:relvars:orientations} for an illustration.
For three orientations $a,b,c$, we now consider the triple
\[
	\langle\
	\ms{Or}(\sphericalangle(b,a)),\;
	\ms{Or}(\sphericalangle(c,b)),\;
	\ms{Or}(\sphericalangle(c,a)) \ \rangle.
\]
Of all $4^3$ triples over $\{\ms{e},\ms{l},\ms{o},\ms{r}\}$,
only 24 combinations are geometrically possible.
We denote this set by $\ms{Cyc}$.
Fig.~\ref{fig:relvars:cyc} shows three of its elements.

Such cyclic ordering information can be expressed within the
relation variable approach in an array $\mi{CycRel}$,
which in particular is ternary.
We have thus a relation variable
\[
	\mi{CycRel}[a,b,c] \in \ms{Cyc}
	\qquad
	\text{with}\quad
	\ms{Cyc} = \{ \ms{lrl}, \ms{orl}, \ldots, \ms{rle} \}
\]
for every three orientations $a,b,c$.
The integrity constraints here are
\begin{gather*}
	\ms{conv}(\,\mi{CycRel}[a,b,c],\; \mi{CycRel}[a,c,b]\,),\\
	\ms{comp}(\,\mi{CycRel}[a,b,c],\; \mi{CycRel}[a,c,d],\; \mi{CycRel}[a,b,d]\,),
\end{gather*}
and a new constraint
\begin{gather*}
	\ms{rotate}(\,\mi{CycRel}[a,b,c],\; \mi{CycRel}[c,a,b]\,).
\end{gather*}
Details and definitions can be found in \cite{isli:2000:new}.

Working within the relation constraint approach,
Isli and Cohn construct a new algorithm called \emph{s4c}
that enforces $4$-consistency \cite{freuder:1978:synthesizing}
on the ternary relation constraints
that correspond to $\mi{CycRel}$.
They are able to prove that this algorithm decides
consistency, \ie, 2D geometric feasibility,
of fully specified scenarios.
The \emph{s4c} algorithm uses exactly the information that we represent
in the $\ms{conv}$, $\ms{comp}$ and $\ms{rotation}$ constraints.
Consequently, we can conclude that in our relation variable model
these constraints guarantee geometric consistency.

We hypothesise further that \emph{s4c} in the relation constraint model
propagates at most as much information as a GAC-enforcing algorithm
does in our relation variable model.
Intuitively, this should be clear: every possible reduction of a disjunctive constraint
in the relation constraint model corresponds to a domain reduction
of a relation variable in our model.

%---------------------------------------------------------------------

\subsection{Combining Cardinal Direction with Relative Orientation}

Isli \cite{isli:2003:combining,isli:2004:combining}
studies the problem of exchanging information
between a cardinal direction model for pairs of points
as in Section~\ref{section:cardinal-directions},
and a relative orientation model for triples of points,
derived from Freksa and Zimmermann's formalisation~\cite{freksa:1992:utilization}.
This problem is again similar to combining topology and size,
Section~\ref{section:topology-size}.
Isli works with the relation-constraints and
proposes a new algorithm for this integration issue.

We formulate a relation-variable model.
The cardinal direction subproblem can straightforwardly
be expressed in this approach;
we omit the obvious details here.
The relative orientation subproblem leads to a model similar
to that of orientations in the preceding section;
in particular, it is based on a ternary array.
The arrays in the combined model are:
\begin{align*}
	&n \times n \times n \text{ array } \mi{ROrientRel}, \text{ and}\quad\\
	&n \times n \text{ array } \mi{CDirRel},
\end{align*}
if we assume $n$ points.

For linking the two models, Isli~\cite{isli:2004:combining}
devises functions for both directions of the information transfer.
They can be transformed into the two constraints
\[\begin{array}{l}
	\ms{link}_{\mr{CD}\rightarrow\mr{RO}}
	(\,
	\mi{CDirRel}[a,b],\;
	\mi{CDirRel}[b,c],\;
	\mi{ROrientRel}[a,b,c]
	\,),
	\\
	\ms{link}_{\mr{CD}\leftarrow\mr{RO}}
	(\,
	\mi{ROrientRel}[a,b,c],\;
	\mi{CDirRel}[a,b],\;
	\mi{CDirRel}[b,c],\;
	\mi{CDirRel}[a,c]
	\,).
\end{array}\]

For the relation-constraint model
it is necessary to treat the information in
$\ms{link}_{\mr{CD}\rightarrow\mr{RO}}, \ms{link}_{\mr{CD}\leftarrow\mr{RO}}$
as meta-constraints, embedded inside an algorithm
that moreover integrates \emph{s4c}
of \cite{isli:2000:new} and a path-consistency algorithm.

Using relation variables,
it suffices to state the constraints
and provide a generic GAC-enforcing algorithm.
Also, for a given triple of points,
the first constraint $\ms{link}_{\mr{CD}\rightarrow\mr{RO}}$
should just be the restriction of
the second constraint $\ms{link}_{\mr{CD}\leftarrow\mr{RO}}$
in which the variable $\mi{CDirRel}[a,c]$ is projected away.
The former constraint is then redundant, and we just need one constraint
\[
	\ms{link}_{\mr{CD}\&\mr{RO}}
	(\,
	\mi{ROrientRel}[a,b,c],\;
	\mi{CDirRel}[a,b],\;
	\mi{CDirRel}[b,c],\;
	\mi{CDirRel}[a,c]
	\,).
\]

On the grounds that both the relation-variable and the relation-constraint
approach are based on the same semantic information,
for one embedded in an algorithm, for the other in constraints,
we conclude that both accept exactly the same point configuration scenarios.

%=====================================================================

\section{Extensions}

\subsubsection{Variables ranging over spatial objects.}

In the relation-variable model,
spatial objects are denoted by constants.
An \emph{object variable},
whose domain is the set of object constants,
has thus a different meaning than
in the relation-constraint approach.
This issue is best demonstrated by an example.
Suppose we wish to identify two
regions among all given regions such that
	\begin{itemize}
	\item the first is smaller than the second, and
	\item they are disconnected or externally connected.
	\end{itemize}
We use topological and size information,
formalised as in Section~\ref{section:topology-size},
so we have arrays $\mi{SizeRel}$ and $\mi{TopoRel}$
recording the qualitative relations.
Let $\Regions$ be the set of the $n$ region constants.
We define the
\[
	\text{region variables } x_1, x_2
\]
whose domain is the set $\Regions$,
and constrain them by
\begin{align}
	&\mi{SizeRel}[x_1, x_2] = (<),
	\label{eq:regionarrays-example-1}\tag{$C_1$}
	\\
	&\mi{TopoRel}[x_1, x_2] \in \{\ms{DC},\ms{EC}\}.
	\label{eq:regionarrays-example-2}\tag{$C_2$}
\end{align}
$C_1$ is a constraint on the variables $x_1,x_2$
and on all size relation variables in the array $\mi{SizeRel}$.
Namely, region constants $r_1,r_2 \in \Regions$ must be assigned to $x_1,x_2$
such that the size relation variable $\mi{SizeRel}[r_1,r_2]$
is assigned a `$<$'.

We call such constraints, in which arrays are indexed
by variables instead of constants, \emph{array constraints}.
They are a generalisation of the better-known \textsf{element}
constraint, which corresponds to a one-dimensional array indexed
by a variable.
Constraint propagation to establish GAC for array constraints
is studied in \cite{brand:2001:constraint2}.
The constraint programming system ILOG Solver~\cite{ilog:2001:solver}
accepts and propagates array constraints.

\subsubsection{Reasoning about spatial change.}

It is not difficult to augment a relation-variable model
with temporal information.  It suffices to add
a new time index to each array of qualitative relations,
and to link the new time-annotated scenarios appropriately.
We extend $\Rel$ from a binary to a ternary array such that
\[
	\Rel[a, b, t]
\]
is a variable specifying the relation between the spatial objects
$a$ and $b$ at time $t$.
Suppose we view time as linear and discrete, such that only atomic
relational changes can occur between subsequent time points.
We can specify these atomic changes (the so-called conceptual neighbourhood)
by pairs of qualitative relations and define accordingly a new binary constraint $\ms{neighbour}$.
For example, the pair $(\ms{DC}, \ms{EC})$ in the constraint
$\ms{neighbour}_{\mr{Topo}}$ indicates that the topological relation
\emph{disconnected} between two regions may change in one time step
to \emph{externally connected}.
The $\ms{neighbour}$ constraint is then stated on
all variable pairs $(\Rel[a, b, t], \Rel[a, b, t'])$
where $t$ directly precedes $t'$ temporally.

%=====================================================================

\section{Summary}

We have presented an alternative formulation
of qualitative spatial reasoning problems as
constraint satisfaction problems.
Contrary to the conventional approach,
we model qualitative relations as variables.
Uncertain relational information is naturally expressed by
variables with domains; consistency of this information
is naturally expressed by static constraints.
The propagation of these constraints is a well-understood issue
in research on constraint programming, and
corresponding generic algorithms are provided by many
constraint solving systems.

While the principle of the relation-variable approach is not new,
the advantages of applying it to QSR,
especially for integration tasks,
have so far very rarely been realised.
We have argued that several algorithms that are custom-designed for integrating
spatial aspects
become unnecessary if a relation-variable model
and a generic GAC-establishing constraint propagation algorithm
is used:
the \textsc{Bipath-consistency} algorithm of \cite{gerevini:2002:combining},
the \emph{s4c} algorithm of \cite{isli:2000:new},
the algorithm combining \emph{s4c} and a path-consistency algorithm
of \cite{isli:2004:combining}.
We have shown how the relation-variable approach can
accommodate composite qualitative relations
as investigated in \cite{cicerone:2004:cardinal,skiadopoulos:2001:composing}
with the help of set variables and constraints.
We have indicated that extending or combining a relation-variable model
often consists mainly in defining appropriate constraints,
contrary to what is the case in the relation-constraint approach
where new algorithms must be designed.

Finally, we remark that the strictly declarative model
that results from using relation-variables can be solved
by any sufficiently expressive solver of CSPs.
This includes typical CP systems based on search and propagation,
but also for example solvers based on local search.

%=====================================================================

\bibliographystyle{plain}

\begin{thebibliography}{10}

\bibitem{apt:2003:principles}
K.~R. Apt.
\newblock {\em Principles of Constraint Programming}.
\newblock Cambridge University Press, 2003.

\bibitem{bennett:1998:determining}
B.~Bennett.
\newblock Determining consistency of topological relations.
\newblock {\em Constraints}, 2:213--225, 1998.

\bibitem{bessiere:1997:arc-consistency}
C.~Bessi{\`e}re and J.-C. R{\'e}gin.
\newblock Arc consistency for general constraint networks: preliminary results.
\newblock In {\em Proc. of 15th International Joint Conference on Artificial
  Intelligence ({IJCAI}'97)}, pages 398--404, 1997.

\bibitem{brand:2001:constraint2}
S.~Brand.
\newblock Constraint propagation in presence of arrays.
\newblock In K.~R. Apt, R.~Bart{\'a}k, E.~Monfroy, and F.~Rossi, editors, {\em
  Proc. of 6th Workshop of the ERCIM Working Group on Constraints}, 2001.

\bibitem{cicerone:2004:cardinal}
S.~Cicerone and P.~Di Felice.
\newblock Cardinal directions between spatial objects: the pairwise-consistency
  problem.
\newblock {\em Information Sciences}, 164:165--188, 2004.

\bibitem{cohn:2001:qualitative}
A.~G. Cohn and S.~M. Hazarika.
\newblock Qualitative spatial representation and reasoning: An overview.
\newblock {\em Fundamenta Informaticae}, 46(1-2):1--29, 2001.

\bibitem{cristani:1999:complexity}
M.~Cristani.
\newblock The complexity of reasoning about spatial congruence.
\newblock {\em Journal of Artificial Intelligence Research}, 11:361--390, 1999.

\bibitem{dechter:2003:constraint}
R.~Dechter.
\newblock {\em Constraint Processing}.
\newblock Morgan Kaufmann, 2003.

\bibitem{frank:1992:qualitative}
A.~U. Frank.
\newblock Qualitative spatial reasoning about distance and directions in
  geographic space.
\newblock {\em Journal of Visual Languages and Computing}, 3:343--373, 1992.

\bibitem{freksa:1992:utilization}
C.~Freksa and K.~Zimmermann.
\newblock On the utilization of spatial structures for cognitively plausible
  and efficient reasoning.
\newblock In {\em Proc. of IEEE International Conference on Systems, Man, and
  Cybernetics}, pages 18--21. IEEE, 1992.

\bibitem{freuder:1978:synthesizing}
E.~C. Freuder.
\newblock Synthesizing constraint expressions.
\newblock {\em Communications of the ACM}, 21(11):958--966, 1978.

\bibitem{fruehwirth:2003:essentials}
T.~Fr{\"u}hwirth and S.~Abdennadher.
\newblock {\em Essentials of Constraint Programming}.
\newblock Springer, 2003.

\bibitem{gerevini:2002:combining}
A.~Gerevini and J.~Renz.
\newblock Combining topological and size constraints for spatial reasoning.
\newblock {\em Artificial Intelligence}, 137(1-2):1--42, 2002.

\bibitem{gervet:1997:interval}
C.~Gervet.
\newblock Interval propagation to reason about sets: {D}efinition and
  implementation of a practical language.
\newblock {\em Constraints}, 1(3):191--244, 1997.

\bibitem{goyal:1997:direction}
R.~K. Goyal and M.~J. Egenhofer.
\newblock The direction-relation matrix: A representation of direction
  relations for extended spatial objects.
\newblock In {\em Proc. of UCGIS Annual Assembly and Summer Retreat}, 1997.

\bibitem{ilog:2001:solver}
ILOG S.A.
\newblock {\em {S}olver 5.1 Reference Manual}, 2001.

\bibitem{isli:2003:combining}
A.~Isli.
\newblock Combining cardinal direction relations and relative orientation
  relations in qualitative spatial reasoning.
\newblock Technical report, University of Hamburg, Dept. of Informatics, 2003.

\bibitem{isli:2004:combining}
A.~Isli.
\newblock Combining cardinal direction relations and other orientation
  relations in {QSR}.
\newblock In {\em Proc. of 8th International Symposium on Artificial
  Intelligence and Mathematics ({AI\&M}'04)}, 2004.

\bibitem{isli:2000:new}
A.~Isli and A.~G. Cohn.
\newblock A new approach to cyclic ordering of {2D} orientations using ternary
  relation algebras.
\newblock {\em Artificial Intelligence}, 122(1-2):137--187, 2000.

\bibitem{ligozat:1998:reasoning}
G.~Ligozat.
\newblock Reasoning about cardinal directions.
\newblock {\em Journal of Visual Languages and Computing}, 9(1):23--44, 1998.

\bibitem{mackworth:1977:consistency}
A.~K. Mackworth.
\newblock Consistency in networks of relations.
\newblock {\em Artificial Intelligence}, 8(1):118--126, 1977.

\bibitem{mackworth:1985:complexity}
A.~K. Mackworth and E.~C. Freuder.
\newblock The complexity of some polynomial network algorithms for constraint
  satisfaction problems.
\newblock {\em Artificial Intelligence}, 25:65--74, 1985.

\bibitem{mohr:1988:good}
R.~Mohr and G.~Masini.
\newblock Good old discrete relaxation.
\newblock In Y.~Kodratoff, editor, {\em Proc. of European Conference on
  Artificial Intelligence ({ECAI}'88)}, pages 651--656. Pitman publishers,
  1988.

\bibitem{montanari:1974:networks}
U.~Montanari.
\newblock Networks of constraints: Fundamental properties and applications to
  picture processing.
\newblock {\em Information Science}, 7:95--132, 1974.

\bibitem{randell:1992:spatial}
D.~A. Randell, Z.~Cui, and A.~G. Cohn.
\newblock A spatial logic based on regions and connection.
\newblock In B.~Nebel, C.~Rich, and W.~R. Swartout, editors, {\em Proc. of 2nd
  International Conference on Principles of Knowledge Representation and
  Reasoning ({KR}'92)}, pages 165--176. Morgan Kaufmann, 1992.

\bibitem{skiadopoulos:2001:composing}
S.~Skiadopoulos and M.~Koubarakis.
\newblock Composing cardinal direction relations.
\newblock In C.S. Jensen, M.~Schneider, B.~Seeger, and V.J. Tsotras, editors,
  {\em Proc. of 7th International Symposium on Advances in Spatial and Temporal
  Databases ({SSTD}'01)}, volume 2121 of {\em LNCS}, pages 371--386. Springer,
  2001.

\bibitem{tsang:1987:consistent}
E.~P.~K. Tsang.
\newblock The consistent labeling problem in temporal reasoning.
\newblock In K.~S.~H. Forbus, editor, {\em Proc. of 6th National Conference on
  Artificial Intelligence ({AAAI}'87)}, pages 251--255. AAAI Press, 1987.

\bibitem{wallace:1997:eclipse}
M.~G. Wallace, S.~Novello, and J.~Schimpf.
\newblock {ECLiPSe}: A platform for constraint logic programming.
\newblock {\em ICL Systems Journal}, 12(1):159--200, 1997.

\end{thebibliography}

%=====================================================================

\end{document}